\documentclass[]{spie}  

\usepackage{amsmath,amsfonts,bm}

\def\eqref#1{equation~\ref{#1}}

\def\1{\bm{1}}

\def\rmD{{\mathbf{D}}}

\def\rmI{{\mathbf{I}}}

\def\rmM{{\mathbf{M}}}

\def\rmR{{\mathbf{R}}}

\def\rmT{{\mathbf{T}}}

\def\rmW{{\mathbf{W}}}

\def\vm{{\bm{m}}}

\def\vr{{\bm{r}}}

\DeclareMathAlphabet{\mathsfit}{\encodingdefault}{\sfdefault}{m}{sl}
\SetMathAlphabet{\mathsfit}{bold}{\encodingdefault}{\sfdefault}{bx}{n}

\DeclareMathOperator*{\argmin}{arg\,min}

\usepackage{amsmath,amsfonts,amssymb}
\usepackage{graphicx}
\usepackage[colorlinks=true, allcolors=blue]{hyperref}
\usepackage[capitalize,noabbrev]{cleveref}
\usepackage{algorithm}
\usepackage{algorithmic}
\usepackage{booktabs}
\usepackage{amssymb}
\usepackage{pifont}
\graphicspath{{fig/figs}}

\newcommand{\minisection}[1]{\vspace{.08in}\noindent{\textbf{#1}.}}

\title{TorchResist: Open-Source Differentiable Resist Simulator}

\author[1]{Zixiao Wang}
\author[1]{Jieya Zhou}
\author[1]{Su Zheng}
\author[1]{Shuo Yin}
\author[2]{Kaichao Liang}
\author[2]{\\Shoubo Hu}
\author[2]{Xiao Chen}
\author[1]{Bei Yu}
\affil[1]{
    The Chinese University of Hong Kong,  Hong Kong SAR
}
\affil[2]{
    Huawei, Hong Kong SAR
}

\begin{document} 
\maketitle

\begin{abstract}

Recent decades have witnessed remarkable advancements in artificial intelligence (AI), including large language models (LLMs), image and video generative models, and embodied AI systems. These advancements have led to an explosive increase in the demand for computational power, challenging the limits of Moore’s Law. Optical lithography, a critical technology in semiconductor manufacturing, faces significant challenges due to its high costs. To address this, various lithography simulators have been developed. However, many of these simulators are limited by their inadequate photoresist modeling capabilities. This paper presents \textbf{TorchResist}, an open-source, differentiable photoresist simulator.TorchResist employs an analytical approach to model the photoresist process, functioning as a white-box system with at most twenty interpretable parameters. Leveraging modern differentiable programming techniques and parallel computing on GPUs, TorchResist enables seamless co-optimization with other tools across multiple related tasks. Our experimental results demonstrate that TorchResist achieves superior accuracy and efficiency compared to existing solutions. The source code is publicly available at \href{https://github.com/ShiningSord/TorchResist}{https://github.com/ShiningSord/TorchResist}.

\end{abstract}


\section{Introduction}


The rapidly growing demand for computational density in modern AI applications, such as LLMs \cite{liu2024deepseek,achiam2023gpt,wang2024moreaupruner} and generative AI (GenAI)\cite{rombach2022high,wang2023evaluation} models,4,5 poses significant challenges to the electronics industry. As semiconductor nodes continue to shrink and transistor counts rise, optical lithography\cite{mack2007fundamental}, a critical technology in semiconductor manufacturing, has become indispensable in current integrated circuit (IC) fabrication processes, accounting for approximately 30\% to 40\% of production costs. To reduce these costs, various lithography simulators have been developed\cite{banerjee2013iccad,chen2024open,fuilt,watanabe2017accurate,fuhner2014artificial}. Recent advancements leverage the computational power of GPUs and machine learning (ML)\cite{yu2007true,wang2023diffpattern,watanabe2017accurate,wang2024chatpattern,chen2024ultra} techniques to accelerate simulations. However, their inadequate resist modeling capabilities reduce the overall effectiveness and accuracy of these lithography simulators.

\begin{figure}
    \centering
    \includegraphics[width=\linewidth]{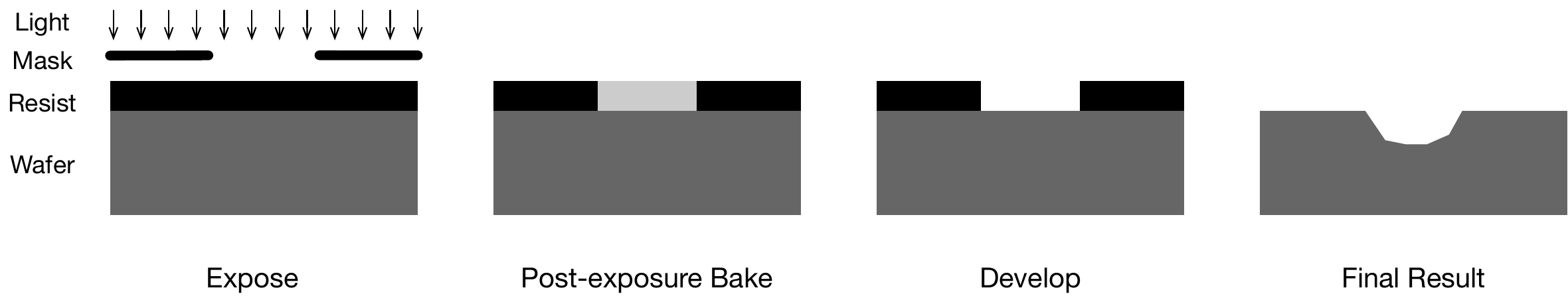}
    \caption{Illustration of a positive resist process.}
    \label{fig:pipeline}
\end{figure}

The basic principle behind the operation of a photoresist is the change in solubility of the resist in a developer upon exposure to light\cite{mack2007fundamental}. The modeling of the resist process\cite{dill1975characterization,mack1987development} commonly involves multiple steps after exposure. These steps include post-exposure bake (PEB), development, and postbake. A simplified process is illustrated in \Cref{fig:pipeline}. During exposure, a strong acid is generated. However, this acid alone does not change the solubility of the resist. During the PEB process, the photogenerated acid catalyzes a reaction that alters the solubility of the polymer resin in the resist. This reaction creates a solubility differential between the exposed and unexposed regions of the resist. Additionally, the PEB process facilitates acid diffusion, which helps to remove standing waves. Development\cite{sethian1996fast} is one of the most critical steps in the photoresist process. In this step, the resist dissolves in the developer, leaving behind the designed patterns on the photoresist. These patterns are used for further pattern transfer. Finally, postbake is applied to harden the resist image. This step ensures that the resist can withstand harsh environments, such as implantation or etching. 

In the early stages of lithography, researchers focused on accurately modeling the physical-chemical processes to predict and control the resist behavior, achieving significant success at larger nodes\cite{thackeray2013pursuit,smith2004lithographic}. However, as lithography progresses to advanced, smaller nodes, strict analytical modeling of the resist process becomes increasingly difficult due to the intricate physical-chemical interactions\cite{thackeray2007chemically} at the nanometer scale and the growing complexity of production techniques\cite{thackeray2011materials}. In recent years, a variety of threshold-based\cite{randall1999variable} and network-based resist models\cite{zach2004neural} have been developed. Threshold-based methods are valued for their simplicity and computational efficiency but are constrained by limited model capacity. In contrast, network-based methods leverage neural networks to model the resist process, iteratively updating the network's parameters using existing aerial-wafer image pairs. While network-based methods excel at predicting data with distributions similar to the training set, their generalization to unseen data remains uncertain, and the resulting models often lack interpretability and explainability.

The core principle of TorchResist is to regularize the resist model using off-the-shelf analytical formulations while delegating parameter calibration to well-established numerical methods, leveraging a given calibration dataset. This strict formulation ensures sufficient model capacity while serving as a strong regularizer, enabling TorchResist to achieve robust generalization on unseen data. Unlike existing network-based methods, which often involve millions of trainable parameters, TorchResist is designed with fewer than twenty interpretable parameters, allowing it to converge efficiently on relatively small datasets. Additionally, by utilizing modern differentiable programming tools and harnessing the parallel computing power of GPUs, TorchResist significantly accelerates both calibration and inference processes.

\minisection{Differentiable Programming and Parallel Computation}
Differentiable programming, supported by tools like TensorFlow\cite{tensorflow2015-whitepaper}, PyTorch\cite{paszke2019pytorch}, and JAX\cite{jax2018github}, enables efficient optimization of computational models through automatic differentiation, with forward and backward methods for gradient computation. Its applications span probabilistic programming, Bayesian inference, robotics, and computational lithography. GPU parallel computing accelerates these tasks by processing multiple operations concurrently, significantly reducing computation time in areas such as semiconductor lithography, where it handles mask and resist results simultaneously, enhancing efficiency in high-performance computing.

\section{Algorithms}

\subsection{Resist Models}

\minisection{Exposition Model} The exposure of the resist is the initial step in the process. As described in previous work \cite{dill1975characterization}, when light passes through the resist without reflection, the Lambert-Beer law can be applied to characterize the optical absorption:
\begin{equation}
    \frac{d\rmI}{dh} = -\rmI \sum_i a_i m_i,
    \label{eq:LBL}
\end{equation}
where $\rmI$ represents the light intensity, $h$ denotes the distance from the resist-air interface, and $a_i$ and $m_i$ are the molar absorption coefficient and molar concentration of the $i$-th component, respectively. We consider three absorbing species: the inhibitor, the base resin, and the reaction products. For a positive photoresist, \Cref{eq:LBL} can be further specified as:
\begin{equation}
    \frac{\partial \rmI(h,t)}{\partial h} = -\rmI(h,t) \left[a_1 \vm_1(h,t) + a_2 \vm_2(h,t) + a_3 \vm_3(h,t)\right],
    \label{eq:depth}
\end{equation}
where $a_1$, $a_2$, and $a_3$ are the molar absorption coefficients of the inhibitor, base resin, and reaction products, respectively. Similarly, $\vm_1$, $\vm_2$, and $\vm_3$ represent the molar concentrations of the inhibitor, base resin, and reaction products. The variables $h$ and $t$ denote the depth in the film and the exposure time, respectively.
 The destruction of inhibitor can be obtained via,
\begin{equation}
    \frac{\partial \vm_1(h,t)}{\partial t} = - \vm_1(h,t)\rmI(h,t)C,\label{eq:time}
\end{equation}
where $C$ is the fractional decay rate of inhibitor per unit intensity.

Assuming the boundary condition of light intensity, the aerial image $\rmR(x,y)$, is given by optical lithography simulation and the lamp intensity is consistent during the exposure, we have,
\begin{equation}
    \rmI(0,t) = \rmR. \label{eq:aerial}
\end{equation}
Considering the initial inhibitor uniformity, resin uniformity and resin does not bleach,
\begin{equation}
\begin{aligned}
    \vm_1(h,0) &= m_{10}; \\
    \vm_2(h,t) &= m_{20}. \label{eq:init}
\end{aligned}
\end{equation}
And reaction product is generated from inhibitor, and the amount of substance is conserved.
\begin{equation}
    \vm_3(h,t) = m_{10} - \vm_1(h,t). \label{eq:amount}
\end{equation}
By substituting \Cref{eq:aerial,eq:init,eq:amount} into \Cref{eq:depth,eq:time}, normalizing, and replacing constants, we obtain:
\begin{equation}
\begin{aligned}
    \frac{\partial \rmI(h,t)}{\partial h} &= -\rmI(h,t)[A\rmM(h,t) + B]; \\
    \frac{\partial \rmI(h,t)}{\partial t} &= -\rmI(h,t) \rmM(h,t)C, \label{eq:expo}
\end{aligned}
\end{equation}
where $\rmM(h,t) = \frac{\vm_1(h,t)}{m_{10}}$ is fractional inhibitor concentration. $A = (a_1-a_3)m_{10}$, $B = (a_2m_{20} + a_3m_{10})$ and $C$ are the constant that should be further calibrated. \Cref{eq:expo} can be further solved with the following initial conditions and boundary conditions,
\begin{equation}
    \begin{aligned}
        \rmM(h,0) &= 1; \\
        \rmM(0,t) &= \exp(-\rmR C t);\\
        \rmI(h,0) &= \rmR\exp[-(A + B)h];\\
        \rmI(0,t) &= \rmR.\label{eq:condition}
    \end{aligned}
\end{equation}

\minisection{Development model} The bulk development model proposed in \cite{mack1987development} can describe the reaction of developer with the resist. Assuming $k_D$ is the rate of diffusion of developer to resist surface, $k_R$ is the rate constant, the rate of development can be described with,
\begin{equation}
    \vr = \frac{k_Dk_R\rmD\vm_3^n}{k_D+k_R\vm_3^n}, \label{eq:speed}
\end{equation}
where $\rmD$ is the bulk developer concentration and we assume $n$ molecules of product $\vm_3$ react with the developer to dissolve a resin molecule. By using \Cref{eq:amount} and the fractional inhibitor concentration $\rmM(h,t)$, \Cref{eq:speed} can be rewritten as,
\begin{equation}
     \vr = \frac{k_D\rmD(1-\rmM)^n}{k_D/k_Rm_{10}^n+(1-\rmM)^n}. 
\end{equation}
As described in \Cref{eq:condition}, when resist unexposed ($t=0$), we have $\rmM=1$ and the rate is zero. When resist completely exposed, we have $\rmM = 0$ and the rate is equal to $r_\text{max}$,
\begin{equation}
    r_\text{max} = \frac{k_D\rmD}{k_D/k_Rm_{10}^n+1}.
\end{equation}
Let $a$ be a constant,
\begin{equation}
    a = k_D/k_Rm_{10}^n.
\end{equation}
The physical meaning of $a$ is an inflection point in the rate curve. By letting,
\begin{equation}
    \frac{d^2\vr}{d\rmM^2} = 0,
\end{equation}
we have,
\begin{equation}
    a = \frac{n+1}{n-1}(1-m_\text{TH})^n,
\end{equation}
where $m_\text{TH}$ is the value of $\rmM$ at the inflection point. By replacing the constant and taking the finite dissolution rate of unexposed resist ($r_\text{min}$) into consideration, the final rate model is,
\begin{equation}
    \vr = r_\text{max}\frac{(a +1)(1-\rmM)^n}{a+(1-\rmM)^n} + r_\text{min},
\end{equation}
where $m_\text{TH}$, $r_\text{max}$, and $r_\text{min}$ should be determined experimentally.

Once the development rate is obtained, the front of developer can be computed by finding the time required to reach each point $\rmT(z,x,y)$, and we have,
\begin{equation}
    \left\vert \nabla\rmT(z,x,y)\right\vert = \frac{1}{\vr(z,x,y)}.
\end{equation}
If we use a simplified model where we only consider the vertical development path, the time can be computed as,
\begin{equation}
    \rmT(z,x,y) = \int_0^h\frac{dh}{\vr(z,x,y)}.
\end{equation}
However, the development path is general not strictly vertical, the fast-marching level-set methods and their variants\cite{osher2001level,sethian1996fast,dai2014fast} can be employed to solve this problem. Given the speed field $\vr$, the time required to reach each point in the field can be computed. And the final development result is the envelop $\rmT(z,x,y) = t_\text{dev}$, where $t_\text{dev}$ is the development time.

In summary, there is a group of parameters should be further decided in TorchResist. Then we decide the values with popular numerical methods.

\subsection{Parameter Optimization}

We have formulated the resist process in previous subsection, and denote it as $f_\theta(\cdot)$, several parameters $\theta$ still wait for further calibration. In our case, the calibration is conducted on a dataset of gray-scale aerial image and binary wafer image pairs $\{(\rmR_i,\rmW_i)\}_{i=1}^N$. The target of the optimization is to minimize the difference of the prediction and target by optimizing the parameter $\theta$ and $\tau$,
\begin{equation}
      \theta, \tau = \argmin_{\theta,\tau} \| \rmW - \Gamma(f_\theta(\rmR) , \tau)\|_0, \label{eq:object}
\end{equation}
where $\Gamma(\cdot, \tau)$ is a threshold function to binaryize the output of resist simulator and $\tau$ is an adjustable threshold. Obviously, directly optimize \Cref{eq:object} is difficult due to the exist of $L0$ norm and threshold function $\Gamma$. 

\minisection{Differentiable Object Functions} To address is issue, a widely applied trick is employing the sigmoid function $\sigma(\cdot)$ and replacing the $L0$-norm with the binary cross-entropy (BCE) between the predicted value and groundtruth.
\begin{gather*}
        \sigma(h) = \frac{e^h}{1+e^h}, \\
        \text{BCE}(h_1, h_2)= -[h_2\log(h_1) + (1-h_2)\log(1-h_1)],
\end{gather*}
where $h_1\in[0,1]$ is the prediction and $h_2\in\{0,1\}$ is the target. And the differentiable object function gives,
\begin{equation}
    \text{loss} = \text{BCE}(\sigma(sf_\theta(\rmR) - s\tau),\rmW), \label{eq:softobject}
\end{equation}
where $s$ is a scale factor to sharpen the boundary of prediction. We should note that every step in the formulation of resist model $f_\theta$ is differentiable, and we implement the process with modern automatic differentiation framework PyTorch\cite{paszke2019pytorch}. Therefore, the optimization of \Cref{eq:softobject} can be easily achieved with popular gradient-decent methods, which we will further detailed in next section.

\section{Numerical Results}

\begin{figure}[t!]
    \centering
    \includegraphics[width=\linewidth]{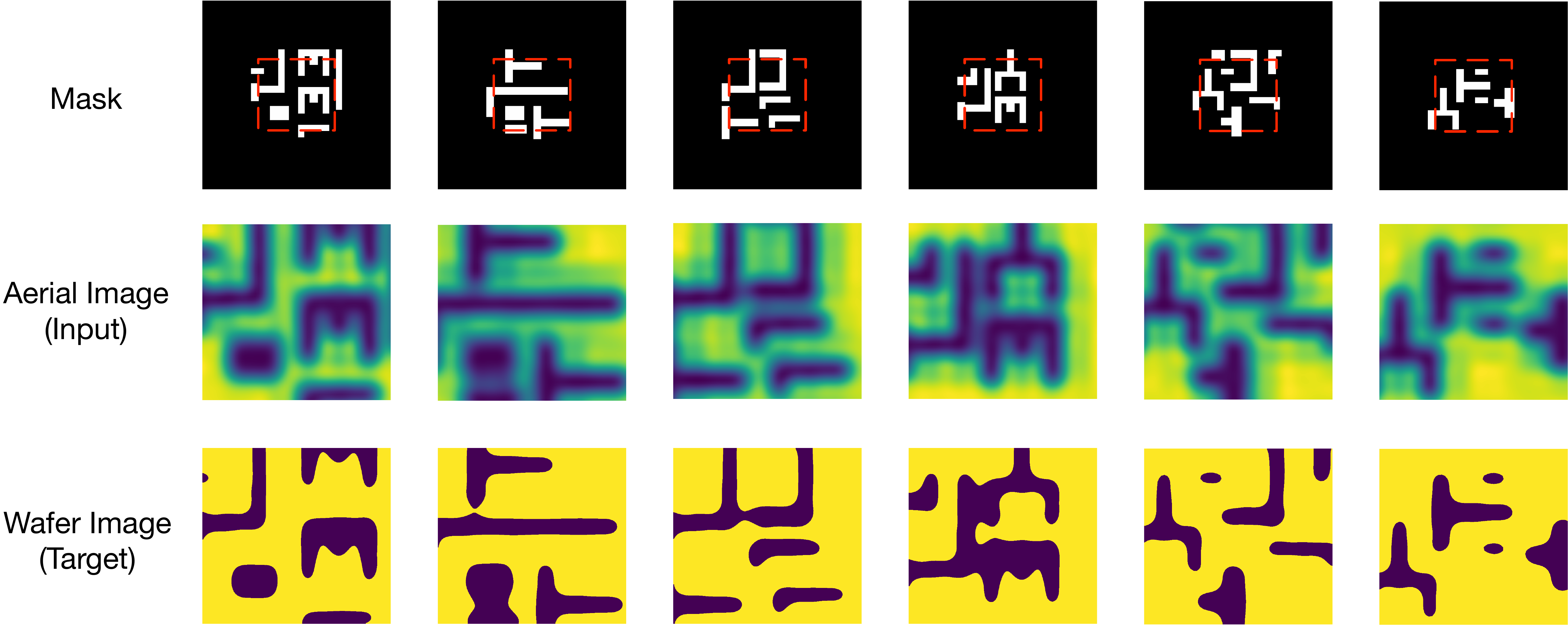}
    \caption{Illustration of the utilized dataset. The region of interest within the mask is highlighted by red dashed lines.}
    \label{fig:dataset}
\end{figure}

\subsection{Experiment Details }

\minisection{Dataset} LithoBench~\cite{zheng2024lithobench} is a well-known layout pattern dataset consisting of 133,496 tiles organized into several subsets. For our work, we focus exclusively on the MetalSet subset, which is generated from the ICCAD-13 benchmark~\cite{banerjee2013iccad} and contains 16,472 tiles, each measuring 2048$\times$2048 nm$^2$. For each tile, we crop the central region with dimensions of 812$\times$756 nm$^2$ and utilize commercial tools to generate both the aerial image and the corresponding resist image. The aerial image is a grayscale representation indicating the intensity distribution in the region of interest, while the resist image is a binary representation that illustrates the resist result. We show several examples in \Cref{fig:dataset}. Both the commercial resist tools and our TorchResist framework ensure that the resolutions of aerial and resist images are consistent at 7$nm$ per pixel during the resist simulation process. We treat the generated aerial images and resist images as the inputs and target outputs of our TorchResist and baseline resist methods. We randomly select 20\% of whole dataset as calibration set and treat the remaining ones as the test set. The calibration/test split is uniform across all resist methods.

\minisection{Parameter Optimization} Some of the parameters are pre-determined based on domain knowledge. For example, in our settings, we set the absorption coefficient of the resist, $\alpha = A + B$, to 6.186$/\text{nm}$, with $A = 0$. The resist thickness is fixed at 75$\,\text{nm}$. The number of $n$ in the development model is set to five for our experiments. Additionally, we always set the boundary sharpness factor $s$ to six.

To calibrate the remaining parameters in TorchResist, we use the popular gradient descent method. We adjust the parameters on the training set for a total of 9 epochs. The learning rate is initialized at 1e-2 and is scaled by a factor of 0.3 after every 3 epochs. The optimizer used is Adam~\cite{kingma2014adam} with $\beta_1, \beta_2 = 0.9, 0.999$. The batch size is set to 16. The entire training process takes approximately 1 hour on a single NVIDIA A100 GPU.

\subsection{Evaluation}

\minisection{Evaluation Metrics} 
After training is completed, we fix the parameters of TorchResist. The predicted developed depth image is first up-sampled to 1$\,\text{nm}$ per pixel and then thresholded by the threshold $\tau$, as explained in \Cref{eq:object}. The up-sampling algorithm used is bilinear interpolation. We compare the predictions of TorchResist on the test set with the corresponding ground truth to evaluate its performance. Quantitatively, we employ three evaluation metrics: Pixel Difference, Edge Placement Error (EPE)-mean, and EPE-max. 

Pixel Difference is the normalized $L_0$-norm between the prediction and the ground truth:
\begin{equation}
     \text{Pixel Difference} = \frac{\| \rmW - \Gamma(f_\theta(\rmR), \tau)\|_0}{\# \text{pixels in total}} \times 100\%.
\end{equation}

EPE estimates the difference between the edges of the ground truth and the predicted edges. EPE-mean is the average EPE value across all edges in a tile, while EPE-max is the maximum EPE value in a tile. The reported values represent the averages across all tiles in the test set.

\begin{table}[t!]
    \centering
    \caption{The performance comparison of the resist model on LithoBench. The lithography model is a commercial tool.}
    \label{tab:acc}
    \resizebox{\textwidth}{!}{
    \begin{tabular}{@{}lccccc@{}}
        \toprule
         Resist Model & Pixel Difference (\%) & EPEMean (nm) & EPEmax (nm) & Differentiable & Depth Simulation \\ 
        \midrule
         Fixed Thres.\cite{banerjee2013iccad} & 0.59 & 1.52 & 4.45 & \ding{55} & \ding{55} \\
         Variable Thres.\cite{randall1999variable} & 0.49 & 1.21 & 3.95 & \ding{55} & \ding{55} \\
         TorchResist & 0.22 & 0.73 & 2.87 & \checkmark & \checkmark \\
        \bottomrule
    \end{tabular}
    }
\end{table}

\begin{figure}[t!]
    \centering
    \includegraphics[width=\linewidth]{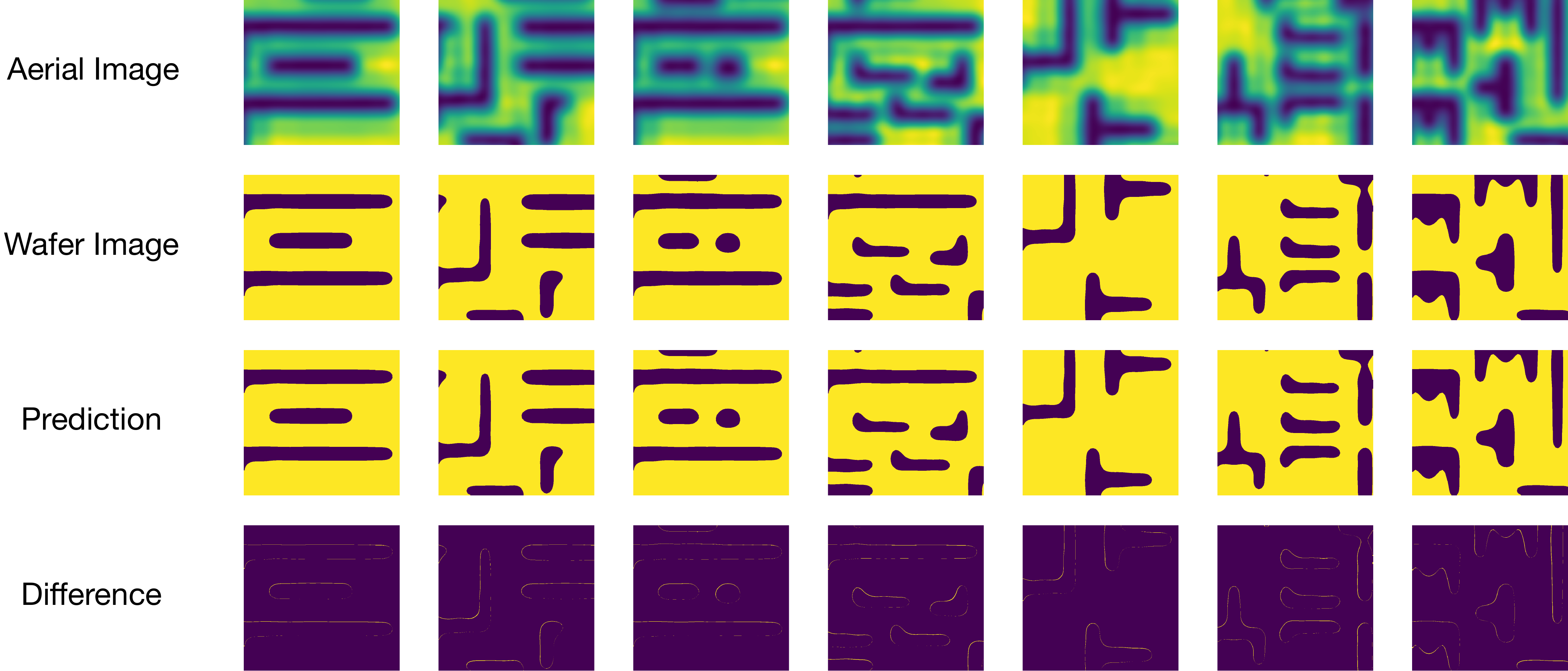}
    \caption{Illustration of the predictions of TorchResist. We also compare the predictions with groundtruth for the reference. The resolution of all the figures are 1 nm/pixel.}
    \label{fig:result}
\end{figure}

\minisection{Baseline Methods} We compare our TorchResist with two baseline methods: Fixed Threshold~\cite{banerjee2013iccad,zheng2024lithobench} and Variable Threshold~\cite{randall1999variable}. The Fixed Threshold method is the simplest resist method, which applies a fixed threshold to the aerial images to obtain the final binary resist results. The Variable Threshold method uses an adaptive threshold to determine the resist result, given by:
\begin{equation}
    \tau_{var} = M_1 + M_2 R_{\text{max}},
\end{equation}
where $M_1$ and $M_2$ are constants to be determined, and $R_{\text{max}}$ is the maximum value of the aerial image in a local region. We fine-tune the constants for both baselines on the training set and evaluate their performance on the test set.

\minisection{Results} We predict the resist values for the aerial images provided by commercial tools using all three resist methods and compare the results in \Cref{tab:acc}. We also show visualizations of the predicted results in \Cref{fig:result}. The results indicate that TorchResist outperforms both threshold-based methods, demonstrating its superior model capacity. Furthermore, TorchResist can simultaneously output both the binary resist and the development depth, which is beneficial for downstream tasks that require 3D simulation results.

\begin{figure}[t]
    \centering
    \includegraphics[width=.8\linewidth]{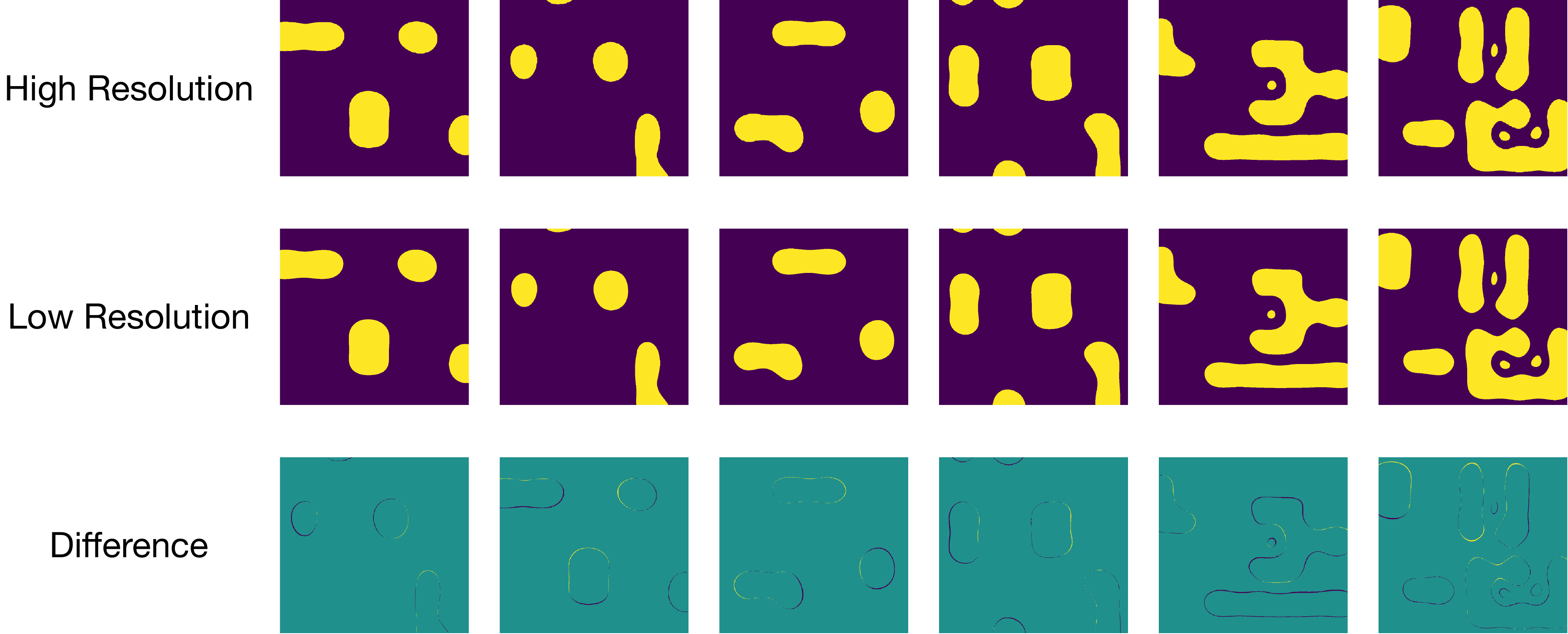}
    \caption{Comparison between results obtained at different resolutions.}
    \label{fig:robust}
\end{figure}

\minisection{Efficiency and Scale Robustness} We also evaluate the efficiency of TorchResist on a single NVIDIA 3090 GPU. We conduct experiments at both 1nm/pixel and 7nm/pixel resolution and report the average processing time for a single mask. The results are summarized in \Cref{tab:time}. We also test the scale robustness of TorchResist by comparing the results at different resolutions. We should note the all the model parameters keep the same under different resolutions. The average pixel difference between them is 0.17\%. Some examples at different resolutions are shown in \Cref{fig:robust}. The result shows the scale robustness of TorchResist, and the inference cost can be largely reduced without an obvious trade-off in precision.

\minisection{Extension on Open-Source Lithography Models} There are two popular open-source lithography simulators that are widely used: FUILT\cite{fuilt} and ICCAD13\cite{banerjee2013iccad}. However, both simulators lack reliable resist modeling, which limits their overall capabilities. To address this limitation and support further research tasks that depend on accurate resist modeling, we introduce two additional variants of TorchResist: TorchResist-F and TorchResist-I. These variants are derived from the outputs of the respective lithography simulators and commercial resist results. 

The evaluation results for these variants are provided in \Cref{tab:other} for reference. It is important to note that the only difference between these variants and the original TorchResist lies in the parameter values, which are adjusted to account for the differences in lithography simulations. Consequently, the evaluation metrics for these variants should not be directly compared with the results presented in the main table.

\begin{table}[t!]
    \centering
    \caption{Comparison of model efficiency, measured by the average time required to process a 2µm $\times$ 2µm patch at different resolutions. Inference is performed using TorchResist on a single NVIDIA 3090 GPU.}
    \label{tab:time}
    \begin{tabular}{c|cc}
    \toprule
        Resolution & Cost Time(s) & Ratio\\
        \midrule
        7 nm/pixel&  0.04  &1.00 \\
         1 nm/pixel&  1.98  &48.96 \\
         \bottomrule
    \end{tabular}
\end{table}

\begin{table}[t!]
    \centering
    \caption{The performance of TorchResist with different open-source lithography models.}
    \label{tab:other}
  
    \begin{tabular}{@{}lcccc@{}}
        \toprule
         Litho Model&Resist Model & Pixel Difference (\%) & EPEMean (nm) & EPEmax (nm) \\ 
        \midrule
         FUILT\cite{fuilt} & TorchResist-F & 1.77&7.03 & 31.91 \\
         ICCAD13\cite{banerjee2013iccad} & TorchResist-I & 3.39 & 10.62 & 59.98\\
        \bottomrule
    \end{tabular}
    
\end{table}
\section{Conclusion}

In this paper, we introduced TorchResist, an open-source, differentiable photoresist simulator designed to address the limitations of existing resist modeling approaches in lithography simulation. By leveraging analytical formulations and modern differentiable programming techniques, TorchResist achieves high accuracy and efficiency compared to traditional methods. Our experimental results on the LithoBench dataset demonstrate that TorchResist significantly reduces pixel difference and edge placement errors, while also providing the capability for 3D depth simulation. Furthermore, the integration of TorchResist with open-source lithography models, such as FUILT and ICCAD13, highlights its versatility and potential for broader applications in semiconductor manufacturing. As an open-source tool, TorchResist is expected to facilitate further research and development in computational lithography, enabling more robust and efficient resist modeling for advanced semiconductor nodes.

\bibliography{report} 
\bibliographystyle{spiebib} 

\end{document}